\documentclass[10pt,twocolumn,letterpaper]{article}

\usepackage[pagenumbers]{cvpr} %

\usepackage[dvipsnames]{xcolor}
\definecolor{cvprblue}{rgb}{0.21,0.49,0.74}
\usepackage[pagebackref,breaklinks,colorlinks,citecolor=cvprblue]{hyperref}
\usepackage{bm}
\usepackage{stmaryrd}
\usepackage{amsmath}
\usepackage{amssymb}
\usepackage{makecell}

\title{Radar Fields: An Extension of Radiance Fields to SAR}

\author{Thibaud Ehret$^{1}$ \quad\quad Roger Marí$^{1}$\quad\quad Dawa Derksen$^{2}$ \quad\quad Nicolas Gasnier$^{2}$ \quad\quad Gabriele Facciolo$^{1}$\\
$^1$Université Paris-Saclay, CNRS, ENS Paris-Saclay, Centre Borelli, France\\
$^2$Centre National d'Études Spatiales, France\\
{\tt\small thibaud.ehret@ens-paris-saclay.fr}
}

\newcommand{\bx}[0]{\bm{x}}
\newcommand{\br}[0]{\bm{r}}
\newcommand{\bv}[0]{\bm{v}}
\newcommand{\bn}[0]{\bm{n}}
\newcommand{\bo}[0]{\bm{o}}

\newcommand{\innerp}[2]{\left\langle #1 \vert #2 \right\rangle}

\DeclareMathOperator{\Tr}{Tr}
\def\figsize{0.20}

\begin{document}
\maketitle

\begin{abstract}
Radiance fields have been a major breakthrough in the field of inverse rendering, novel view synthesis and 3D modeling of complex scenes from multi-view image collections. Since their introduction, it was shown that they could be extended to other modalities such as LiDAR, radio frequencies, X-ray or ultrasound. In this paper, we show that, despite the important difference between optical and synthetic aperture radar (SAR) image formation models, it is possible to extend radiance fields to radar images thus presenting the first ``radar fields''. This allows us to learn surface models using only collections of radar images, similar to how regular radiance fields are learned and with the same computational complexity on average. Thanks to similarities in how both fields are defined, this work also shows a potential for hybrid methods combining both optical and SAR images.
\end{abstract}

\section{Introduction}
\label{sec:intro}

\begin{figure}
    \centering
     \begin{subfigure}[b]{\linewidth}
       \centering \includegraphics[width=\textwidth]{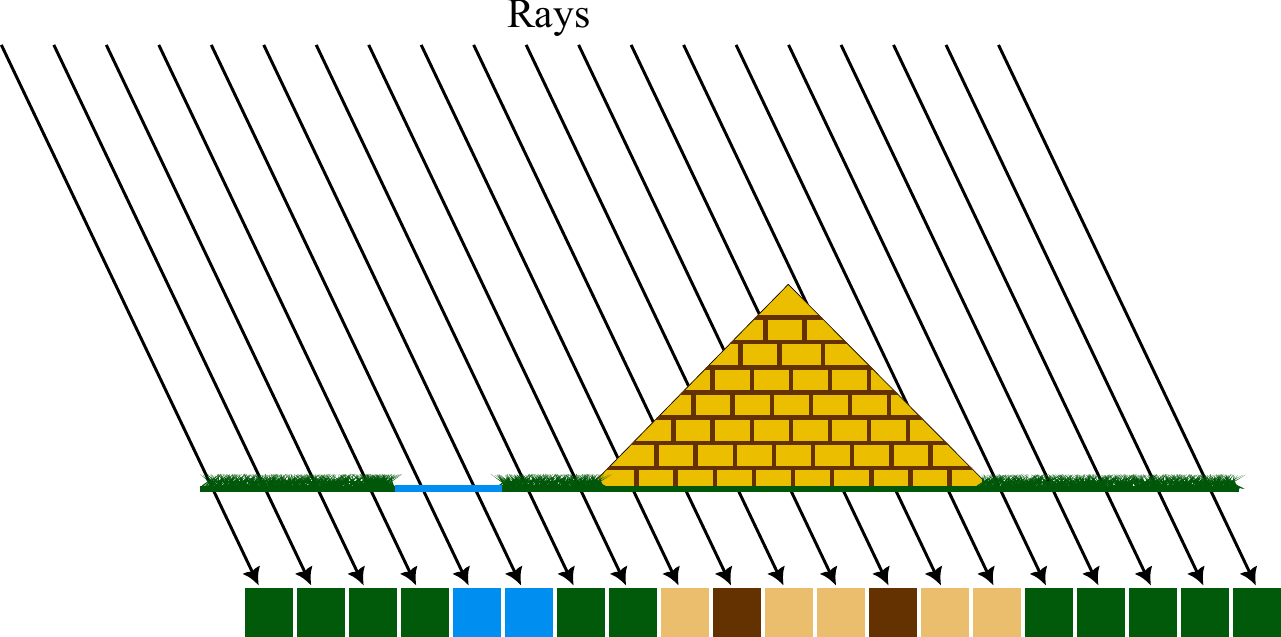}
       \caption{Regular radiance fields}
    \end{subfigure}
    \par\bigskip
     \begin{subfigure}[b]{\linewidth}
       \centering \includegraphics[width=\textwidth]{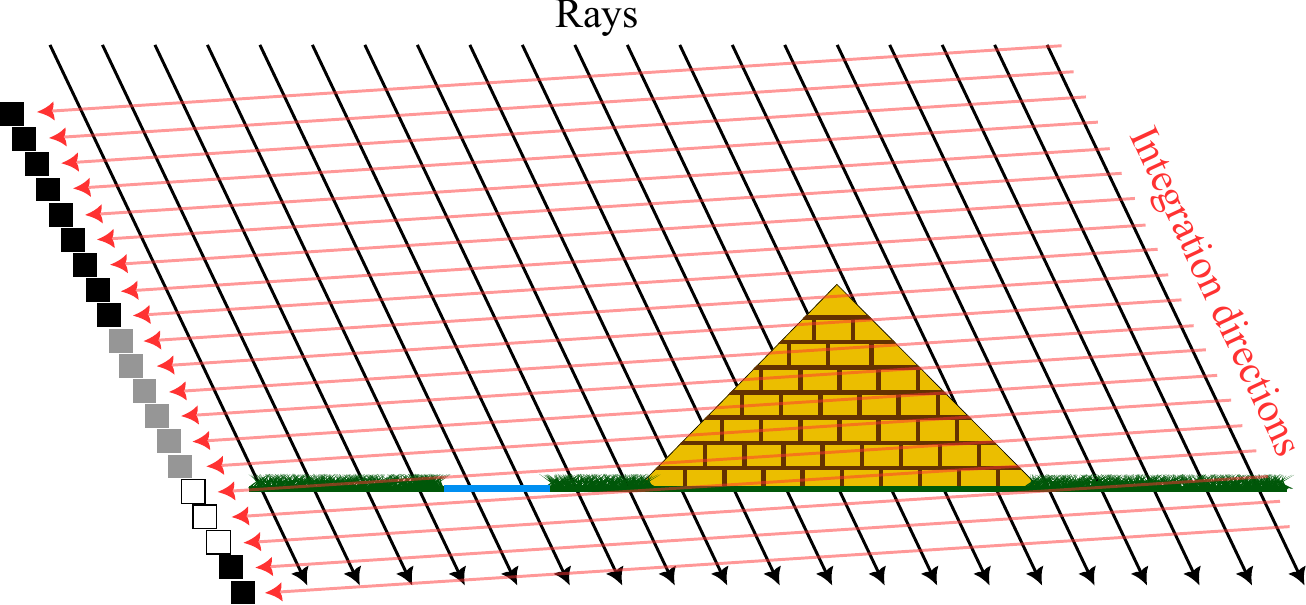}
       \caption{Proposed radar fields}
    \end{subfigure}
    \caption{The difference between the integration process of a classic optical radiance field and the proposed radar field. While the radiance field integrates along each rays (in black), radar fields integrates across rays (in red) in a given azimuth plane. Similarly to radiance fields, radar fields can be optimized from multiple SAR images to recover the underlying surface of the scene.} 
    \label{fig:teaser}
\end{figure}

Many of the latest breakthroughs in the domain of 3D reconstruction have been inspired by Neural Radiance Fields (NeRF) \cite{mildenhall2020nerf}. 
By definition, NeRFs are a variant of Neural Fields that learn a 3D representation of the scene based on the measurement of radiance, given a set of optical images. In the past years, NeRFs have been reshaped, reworked and adapted to various use cases. In particular, their application domain has been extended far beyond computer vision~\cite{gao2022nerf, xie2022neural}. 

This diversity of application domains is largely due to the flexibility of the Neural Fields paradigm. Neural Fields represent a path forward for including physics into the Neural Network learning process, a longstanding aim in AI research. The common goal of all of the Neural Field research works cited in the following section is to learn a high-dimensional representation (e.g. a 3D shape) from a set of low-dimensional data (e.g. 2D images). From a physical point of view, the low-dimensional measurements can be seen as boundary conditions for a set of partial differential equations (PDE) that describe the formation process of the observations. 
In this setting learning amounts to an iterative optimization process to find a high-dimensional representation that simultaneously satisfies both the PDE and the boundary conditions. The learning process generally involves starting with a randomly (or arbitrarily) initialized scene. A set of low-dimensional observations are then computed using a forward ``rendering'' process, based on a discretized form of the PDE. These observations are confronted to the real measurements, yielding a loss. The loss is back-propagated onto the representation to improve its coherence with the observations. This process is repeated until convergence is reached. 

In this work, we focus on using Synthetic Aperture Radar (SAR) data instead of optical images to achieve 3D reconstruction. We introduce Radar Fields, a new framework for learning a volumetric model based on multiple SAR measurements of a scene.
Since the wavelengths used in SAR experience minimal interactions with atmospheric gases or aerosols, the propagation medium is virtually transparent and non-absorbing. Thus, unlike optical sensors, SAR has day-night cloud-penetrating imaging capacities. Unlike interferometric SAR (InSAR), the proposed methodology does not prescribe the geometric configuration of the acquisitions. It could therefore be used with opportunistic acquisitions.
We believe that the main impact of this research will be for global 3D reconstruction, due to the wide-scale availability and high quality of the latest SAR Earth Observation data from missions such as TerraSAR-X~\cite{pitz2010terrasar} or Capella~\cite{farquharson2018capella, stringham2019capella}. The methodology that we present is general and could equally be applied to aerial SAR or to other radar-based measurement systems.

To summarize, our contributions are the following:
\begin{enumerate}
    \item The definition of radar fields, a differentiable rendering framework adapted to radar image synthesis.
    \item A new surface model model definition, inspired by digital surface models, that can easily be used in frameworks like radar fields for remote sensing applications.
    \item Adapting key training techniques, like NeRF random depth sampling or being able to train despite noisy data, to radar fields.
\end{enumerate}

\section{Related works}
\label{sec:related_works}

Our work is not the first to exploit SAR images to produce 3D models. In fact, space-borne SAR sensors are currently used to produce large scale Digital Surface Models (DSM) such as Copernicus DEM~\cite{copernicus2022}. The TanDEM-X mission~\cite{zink2014tandem} uses the principle of radar interferometry to estimate the Earth’s surface topography. The derived Copernicus DEM product describes the elevation of the terrain and built-up land elements with an average error of 2-4 meters. The spatial resolution of the product ranges from 10-30 meters. 

An alternative technique to interferometry, known as SAR tomography~\cite{Tomo2000, shi2020sar}, consists in using stacks of 2D Synthetic Aperture Radar (SAR) images to estimate the surface height with a finer spatial resolution (1-2m). This technique involves first estimating the covariance matrix in each radar cell. In a second step, spectral analysis is used to invert this covariance matrix, resulting in an approximation of the surface height distribution. The resulting DSM achieves a mean altitude error ranging from 1-2 meters in urban areas. However, SAR Tomography commonly suffers from missing parts in urban surfaces, particularly in flat areas such as ground or rooftops. Moreover, the spatial heterogeneity of the height distribution near abrupt altitude transitions can result in a blurry reconstruction~\cite{rambour2019urban}. 

Other works propose deep learning-based single-image height estimation methods using a Convolutional Neural Network encoder-decoder architecture~\cite{recla2022deep, recla2023relative}, taking a SAR image as input and directly predicting depth maps. %
These methods provide fast inference times, but the training requires large amounts of ground-truth 3D data synchronized with multi-view SAR acquisitions to generalize to different viewing angles or landscapes.

Neural fields have seen a number of applications in optical-based satellite imagery, starting with the seminal work S-NeRF~\cite{derksen2021shadow} followed by Sat-NeRF~\cite{mari2022satnerf}, and %
recently, EO-NeRF~\cite{mari2023multi}. S-NeRF initially tackled the problem of varying lighting conditions; Sat-NeRF introduced a latent transient vector to disentangle temporary objects such as cars from the static scene, and EO-NeRF achieved state-of-the-art 3D reconstruction quality by explicitly modeling the shadow effects  and learning per-image color corrections. Another recent work, SparseSat-NeRF~\cite{zhang2023sparsesat}, reduced the need for a large number of input images using depth priors based on stereo-photogrammetry. Season-NeRF~\cite{gableman2023incorporating} models seasonal variations to enable temporal interpolation between different acquisition dates. Sat-Mesh~\cite{qu2023sat} replaced the NeRF model with an implicit surface  representation, yielding impressive visual DSM quality, but not disentangling the transient objects or generalizing to unseen solar angles. RS-NeRF~\cite{xie2023remote} and SatensoRF~\cite{zhang2023fast} have successfully reduced the optimization time, but like Sat-Mesh do not generalize to unseen solar angles. In the domain of aerial images, issues such as large-scale data \cite{xu2023grid, turki2022mega} and image co-registration \cite{chen2023dreg} have also been addressed with NeRF approaches.

Since the arrival of NeRFs, researchers in various fields have also successfully applied the Neural Fields paradigm to entirely different data than optical images. For example, in the domain of radiative transfer and light modeling, these include applications to LiDAR~\cite{huang2023neural, tao2023lidar}, Time-of-flight cameras \cite{attal2021torf} and non-line of sight cameras~\cite{shen2021non}. Impressive results have also been achieved using event based cameras, which detect a difference in pixel intensity rather than the absolute pixel intensity~\cite{rudnev2023eventnerf, low2023robust, hwang2023ev, klenk2023nerf}. In other regions of the electromagnetic spectrum, applications have risen for 3D reconstruction based on X-rays~\cite{chen2023cunerf, corona2022mednerf} and Radio Frequency~\cite{zhao2023nerf}. The equations of light propagation being similar in nature to the equations of sound propagation, the same paradigm can be adapted to the acoustic setting~\cite{chen2023novel}, with applications in ultra-sound~\cite{li20213d, wysocki2023ultra} and sonar~\cite{reed2021implicit}. Finally, the principle of implicit learning has also yielded exploratory works in geodesy~\cite{izzo2021geodesy} and gravitational lensing~\cite{levis2022gravitationally}.

\section{Introduction to radiance fields}
\label{sec:rf_fundamentals}

NeRF \cite{mildenhall2020nerf} represents a static scene as a continuous volumetric function $\mathcal{F}$. This function can, for example, be encoded by a fully-connected neural network (MLP)~\cite{mildenhall2020nerf}, a voxel decomposition~\cite{karnewar2022relu} or a tensor interpolation~\cite{chen2022tensorf}. $\mathcal{F}$ predicts the emitted RGB color~$\mathbf{c} = (r, g, b)$ and a non-negative scalar volume density~$\sigma$ at a 3D point $\mathbf{x} = (x, y, z)$ of the scene seen from a viewing direction $\mathbf{d} = (d_x, d_y, d_z)$, \ie
\begin{equation}
    \mathcal{F}:(\mathbf{x}, \mathbf{d}) \mapsto (\mathbf{c}, \sigma).
    \label{eq:classic_rf_inputs_outputs}
\end{equation}
Multi-view consistency is encouraged by restricting the network to predict the volume density $\sigma$ based only on the spatial coordinates $\mathbf{x}$, while allowing the color $\mathbf{c}$ to be predicted as a function of both $\mathbf{x}$ and and other view-dependent features, such as the viewing direction $\mathbf{d}$.

Given a set of input views and their camera poses, $\mathcal{F}$ is optimized by rendering the color of individual rays traced across the scene and intersecting the known pixels. Each ray $\textbf{r}$ is defined by a point of origin $\mathbf{o}$ and a direction vector $\mathbf{d}$. The color $\mathbf{c}(\mathbf{r})$ of a ray $\mathbf{r}(t) = \mathbf{o} + t\mathbf{d}$ is computed as
\begin{equation}
    \mathbf{c}(\mathbf{r}) = \sum_{i=1}^{N}T_i\alpha_i\mathbf{c}_i.
    \label{eq:rf_color_rendering}
\end{equation}
The rendered color $\mathbf{c}(\mathbf{r})$ results from the weighted integration of the colors $\mathbf{c}_i$ predicted at $N$ different points sampled along the ray $\mathbf{r}$.
Each point $\mathbf{x}_i$ in $\mathbf{r}$ is obtained as $\mathbf{x}_i = \mathbf{o} + t_i\mathbf{d}$, where $t_i$ is the depth step.

{
The contribution of each point to the rendering equation~\eqref{eq:rf_color_rendering} follows a light transmittance model based on the opacity $\alpha_i$ and transmittance $T_i$ values, which  follow the geometry defined by the density $\sigma$ %
\begin{equation}
    \alpha_i = 1 - \exp(-\sigma_i(t_{i+1} - t_i));  \quad 
     T_i = \prod_{j=1}^{i-1} \left( 1 - \alpha_j \right).
     \label{eq:opacity_and_transmittance}
\end{equation}
The opacity $\alpha_i$ increases with $\sigma_i$ and is the probability that a point belongs to a non-transparent surface. The transmittance $T_i$ is
the probability that light arrives without hitting previous opaque points in the ray $\mathbf{r}$.
}

Given \eqref{eq:opacity_and_transmittance}, the depth $d(\mathbf{r})$ observed in the direction of a ray $\mathbf{r}$ can be rendered in a similar manner to \eqref{eq:rf_color_rendering} \cite{deng2021depth,roessle2021dense} as
\begin{equation}
    d(\mathbf{r}) = \sum_{i=1}^{N}T_i\alpha_it_i.
    \label{eq:nerf_depth_rendering}
\end{equation}
The radiance field is optimized by minimizing the mean squared error (MSE) between the rendered color and the %
{actual color of the input views:}
\begin{equation}
    \sum_{\mathbf{r} \in \mathcal{R}} \|  \mathbf{c}(\mathbf{r}) - \mathbf{c}_{\text{GT}}(\mathbf{r}) \|_2^2,
    \label{eq:nerf_classic_loss} 
\end{equation}
where $\mathbf{c}_{\text{GT}}(\mathbf{r})$ is the observed color of the pixel intersected by the ray $\mathbf{r}$, and $\mathbf{c}(\mathbf{r})$ is the color predicted by the radiance field using \eqref{eq:rf_color_rendering}. $\mathcal{R}$ is the set of rays in each input batch.

\paragraph{Surface models.} The concept of radiance fields has been extended to surface models~\cite{yariv2021volume,wang2021neus,darmon2022improving, fu2022geo, li2023neuralangelo}. Instead of considering the scene as continuous volumetric function $\mathcal{F}$, the scene is represented as a surface $\mathcal{S}$. This surface is characterized by an implicit function $\mathcal{F}$ such that  
\begin{equation}
    \mathcal{S} = \left\{\bx\in \mathbb{R}^3~|~\mathcal{F}(\bx) = 0\right\}.
    \label{eq:implicit_surface}
\end{equation}
Denoting $d$ the minimum distance between $\bx$ and $\mathcal{S}$, $\mathcal{F}$ is defined as a signed distance function (SDF) with
\begin{equation}
    \mathcal{F}(\bx) = \begin{cases}
        d(\bx, \mathcal{S}) & \mbox{if }\, \bx \text{ is outside } \mathcal{S},\\
        -d(\bx, \mathcal{S}) & \mbox{if }\, \bx \text{ is inside } \mathcal{S}.
\end{cases}
    \label{eq:def_sdf}
\end{equation}
This property is enforced %
{using} an additional loss term to enforce the Eikonal property of the SDF 
\begin{equation}
    \mathcal{L}_{Eikonal} = \sum_{x \in \Omega} \left(\|\nabla \mathcal{F}(\bx)\|_2^2 - 1\right)^2
    \label{eq:eikonal_loss}
\end{equation}
with $\Omega$ containing points sampled near the surface as well as uniformly in the entire volume.

The {optimization} of a surface based model is similar to {that} of a regular radiance field. The surface model is transformed into a volumetric model using a function that produces a pseudo-density from the SDF value at a given point. For example, in VolSDF~\cite{yariv2021volume}, the transformation function $\Psi$ is defined with the cumulative distribution function of the Laplace function
\begin{equation}
\Psi(d) = \begin{cases} 
    \frac{1}{2} \exp\left(-\frac{d}{\beta}\right) & \text{for } d\geqslant 0 \\
    1-\frac{1}{2}\exp\left(\frac{d}{\beta}\right) & \text{for } d<0,
\end{cases}
\label{eq:laplace}
\end{equation}
where $\alpha$ and $\beta$ %
{are optimized with} the rest of the model. %
{The rendering operation remains unchanged, as~\eqref{eq:rf_color_rendering} and~\eqref{eq:opacity_and_transmittance} use the pseudo-density}  
\begin{equation}
    \sigma_i = \Psi\left(\mathcal{F}\left(\bx_i\right)\right).
\end{equation}

{Representing the geometry with an implicit function $\mathcal{F}$ opens the door to exploit additional information, such as the normal of the surface, \ie} $n(\bx) = \nabla \mathcal{F}(\bx)$ when  $\nabla \mathcal{F}(\bx) \neq \textbf{0}$, which is  prevented by the Eikonal constraint.

\section{Radar fields}
\label{sec:radar_fields}

The models presented in Section~\ref{sec:rf_fundamentals} have been proposed for optical images. We propose here to extend
the concept of radiance fields to SAR images.

\begin{figure}
    \centering
    \includegraphics[width=\linewidth]{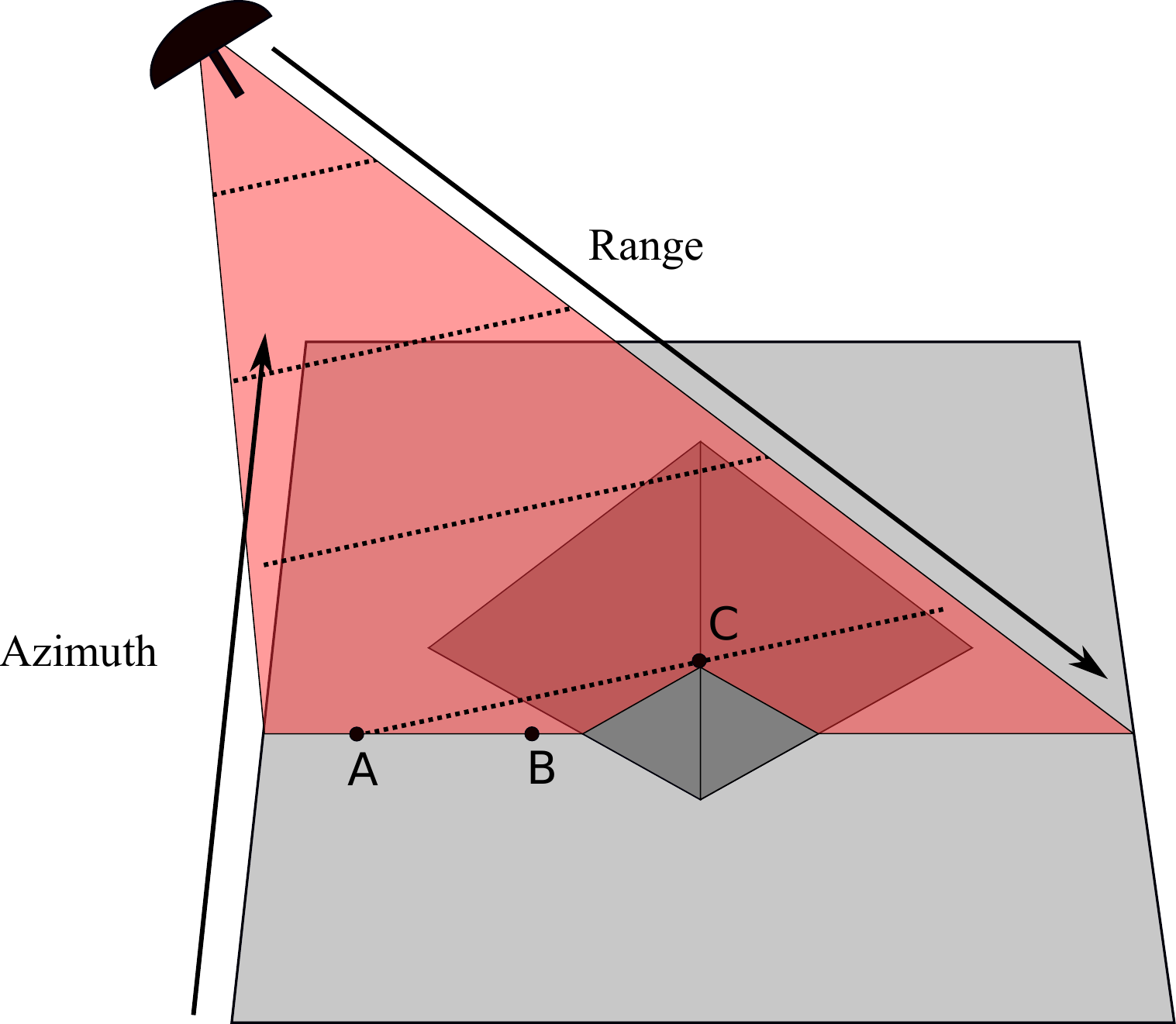}
    \caption{Acquisition process of a SAR image: An electromagnetic pulse is sent from the antenna and Points A and C, at the same distance from the sensor, are projected to the same pixel while B  is projected on a different pixel.}
    \label{fig:acquisition_process}
\end{figure}

\paragraph{Introduction to the SAR image formation model.} A SAR imaging device is an active system. It is usually composed of  an antenna mounted on a moving spaceborne or airborne platform that emits electromagnetic pulses in a side-looking geometry and then receives echoes that are backscattered by the ground surface. 
The satellite flight direction is called along-track or azimuth, while the direction of the beam is called across-track or range. 
For each azimuth pulse emitted, the satellite receives the echoes, demodulates and samples them, orders them by range, and stores them as complex numbers separating the amplitude and phase components of the signal. The amplitude is related to the ground reflectivity and angle, while the phase contains information about the satellite-target path delay (modulo the wavelength).
The resulting image is said to be the raw format. Because of the large footprint of the electromagnetic beam and the length of the pulse, in this format, the response from a single scatterer on the ground is spread over many pixels and the spatial resolution is very low. The resolution is largely improved through a process known as focusing~\cite{Cumming2005}, which yields a complex-valued image equivalent to employing short pulses (for range) and a narrow beam perpendicular to the motion of the satellite known as zero-doppler plane (corresponding to azimuth). 
Unlike optical images, SAR images represent the distance to reflectors in the range direction. If two reflectors are at the same distance from the satellite (in the same azimuth plane), the response will be the sum of both, resulting in foreshortening, layover and shadow effects.
Luckily, this range image can be analogously interpreted as a projected view of the scene from a direction perpendicular to the incidence angle and illuminated by the radar pulse, where objects appear as if they are transparent. The SAR acquisition process is illustrated in Figure~\ref{fig:acquisition_process}.

\paragraph{Definition of radar fields.} %
Using the notations from Section~\ref{sec:rf_fundamentals}, we want to learn the function $\mathcal{F}$ corresponding to the surface model of the scene using a collection of SAR images.
In this section, we derive the model for a given azimuth plane. This is not a limitation since  the image is equivalent to a zero-doppler acquisition for each azimuth plane as mentioned in the previous paragraph, except in the case of double or multiple bounces, which we discuss in Section~\ref{sec:discussion}.

Consider a set of rays $(\br_j)$, with respective origin and direction $(\bo_j, \bv_j)$, in the given azimuth plane and $(d_i)_{i \in \llbracket 1, N \rrbracket}$ the distance sampling associated to these rays. Let $s_{\br_j}(d_i)$ be the sensed signal associated to $\br_j$ at distance $d_i$. Following Section~\ref{sec:rf_fundamentals}, this yields

\begin{equation}
    s_{\br_j}(d_i) = -T_{i,j} \alpha_{i,j} \innerp{\bv_j}{\bn_{i,j}}^\theta,
    \label{eq:radar_signal}
\end{equation}
where the definition of $T_{i,j}$ and $\alpha_{i,j}$, for a fixed $j$, is the same as in~\eqref{eq:opacity_and_transmittance}, $\bn_{ij} = \nabla \mathcal{F}(\bo_j + d_i \bv_j)$, and where the coefficient $\theta$ is used to model the specularity of the reflection~\cite{balz2009hybrid,auer20113d}. 
Indeed, in the case of SAR images, there is no color: the intensity of the reflected signal is characterized by the surface. Traditionally in SAR simulation~\cite{balz2009hybrid,auer20113d}, the reflected signal $s$ is given by
\begin{equation}
    s = -\innerp{\bv}{\bn}^\theta
    \label{eq:lambertian_reflectance}
\end{equation}
with $\bv$ the incidence direction and $\bn$ the normal to the surface. Often a purely Lambertian model, corresponding to $\theta=1$, is assumed. Larger values of $\theta$ are used to represent specular surfaces. The parameter $\theta$ can be learned at the same time as the location of the surface.

As described in the previous paragraph and represented in Figure~\ref{fig:teaser}, the integration is performed \textit{across} rays for SAR images and not \textit{along} rays as in optical images: for a given azimuth, reflectors that are at the same distance are projected into the same pixel. {To satisfy this condition, the rendering operation}~\eqref{eq:rf_color_rendering} must be replaced by 
\begin{equation}
    s(d_i) = \sum_{\br_j} s_{\br_j}(d_i) = - \sum_{\br_j} T_{i,j} \sigma_{i,j} \innerp{\bv_j}{\bn_{i,j}}^\theta.
    \label{eq:radar_integration}
\end{equation}
Note how the role of indices $i$ and $j$ are inverted between Eq.~\eqref{eq:rf_color_rendering} and Eq.~\eqref{eq:radar_integration} and how $\mathbf{c}$ depends on the ray while $s$ depends only the distance.
While the position of rays is well defined by the sensor configuration in the optical case (see Sec.~\ref{sec:rf_fundamentals}), this is not the case for SAR. Conversely, the sampling of the $\bx_{i,j}$ is perfectly defined in the SAR case -- following the range sampling -- but is flexible in the optical case. Finally, computing the results for all the azimuth planes gives the complete SAR image.

\paragraph{Differences between radiance and radar fields.} While radiance and radar fields are conceptually similar, we list some important differences here. They are also summarized in Table~\ref{tab:differences_radfields}. The first difference is that to generate a single SAR pixel, it is necessary to compute all rays corresponding to its azimuth plane. Thankfully, computing a single pixel provides the result for the entire range line at the same time. This means that while rendering a single pixel requires more computations, the overall amount of computations necessary to learn a radar field remains the same, on average, as the one needed to learn a radiance field.  
In practice the only difference is how rays are batched. While there is no requirement in the radiance field case, it is necessary to batch all rays of a given azimuth plane together during computation for radar fields. Note that with additional assumptions on the scene, it is possible to slightly relax this requirement. Indeed, if an approximate position of the surface is known \textit{a priori}, then it would be possible to split the distance sampling into multiple smaller batches that can be computed separately. Note that no such assumption is made for the experiments presented in Section~\ref{sec:experiments}.

A key technique to improve the performance in NeRF~\cite{mildenhall2020nerf} is to introduce some noise when sampling along rays instead of using a fixed uniform sampling.
This is not possible with a radar field because the sampling step along rays is defined by the range sampling. It is however possible to transpose this idea to ray sampling since, as shown in Eq.~\eqref{eq:radar_integration}, the final signal does not require a specific ray position.
as long as $\bo_j$ stays in its corresponding azimuth plane and that it does not impact the range sampling. If we denote by $\bm{w}_j$ the vector orthogonal to $\bv_j$ inside the azimuth plane, then the perturbed ray origin is defined by
\begin{equation}
    \tilde{\bo}_j = \bo_j + n \bm{w}_j \text{ with } n \sim \mathcal{N}(0,1).
    \label{eq:origin_sampling}
\end{equation}

While optical images have a good SNR in normal illumination conditions, SAR images are fundamentally noisy.  They exhibit speckle noise, which has been studied and shown to follow a complex Wishart distribution \cite{Goodman:76,Goodman63}. The distribution $p$ of the noisy sample covariance matrix $\bm{C}$ of dimension $d$ corresponding to the noise-free covariance $\bm{\Sigma}$ in a $L$ looks configuration is
\begin{equation}
    p(\bm{C}) = \frac{L^{L d} |\bm{C}|^{L-d}}{\Gamma_d(L) |\bm{\Sigma}|^L} \exp\left(-L \Tr\left(\bm{\Sigma}^{-1} \bm{C}\right)\right),    
    \label{eq:noise}
\end{equation}
with $\Gamma_d(L) = \pi^{d(d-1)/2} \prod_{k=1}^d \Gamma(L - k + 1)$.
When working with single-channel single-look intensity images, the sample covariance matrix reduces to an intensity $I$ and this model can be simplified into a multiplicative speckle noise with regard to the noise-free reflectance $R$, such that
\begin{equation}
    I = n\times R \quad\quad \text{with} \quad\quad n \sim \Gamma(1, 1). 
    \label{eq:simplified_noise}
\end{equation}
Learning the surface despite the noise is not a problem in practice. Indeed, comparing the noiseless generated view to the noisy samples is sufficient. This is similar to Noise2Noise~\cite{lehtinen2018noise2noise} that shows that it is possible to learn a denoising network despite having only noisy samples. SAR2SAR~\cite{dalsasso2021sar2sar} showed that this idea can also be applied to SAR data and \cite{ehret2019joint} that it is possible to learn the noiseless images of a scene using a single burst of images, even when the noise has a bias. %

\begin{table}
    \centering
  {\small
  \begin{tabular}{l|cc}
    \toprule
    &  Radiance field   &  Radar Field \\
    \midrule
    Complexity & \makecell{height $\times$ width \\$\times$ samples} & \makecell{azimuth $\times$ range \\$\times$ samples (on average)} \\
    \midrule
    Integration & Along the ray & Across the ray\\
    \midrule
    \makecell[l]{Distance\\sampling} & \makecell{Variable\\(along a ray)} & \makecell{Fixed\\(defined by data)} \\
    \midrule
    \makecell[l]{Ray\\sampling} & \makecell{Fixed\\(defined by data)} &  \makecell{Variable\\(in a given azimuth plane)}\\
    \midrule
    Noise & Noise-free & Follows Eq.~\ref{eq:simplified_noise} \\

    \bottomrule
  \end{tabular}
  }
  \caption{Main differences between regular radiance fields and the proposed radar fields.}
  \label{tab:differences_radfields}
\end{table}

\begin{figure*}
    \centering
    \includegraphics[width=\figsize\linewidth]{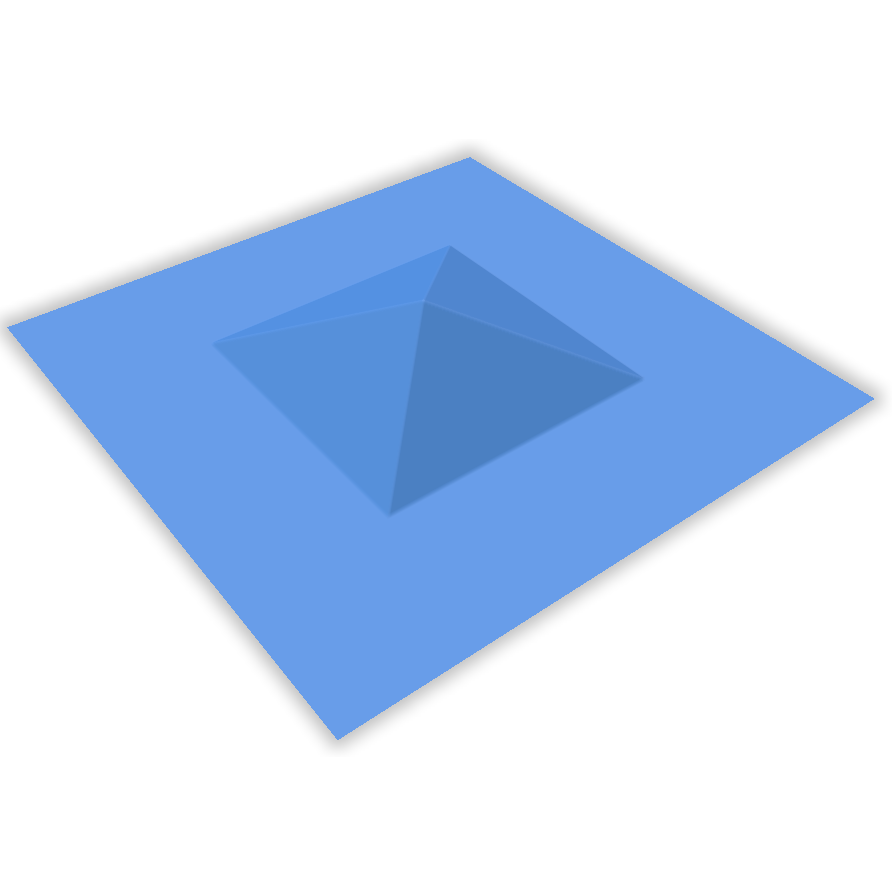}
    \includegraphics[width=\figsize\linewidth]{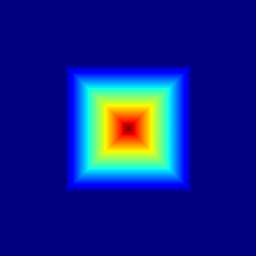}
    \includegraphics[width=\figsize\linewidth]{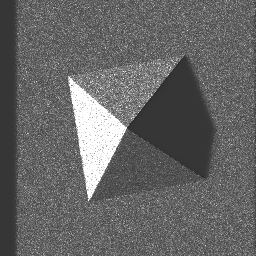}
    \includegraphics[width=\figsize\linewidth]{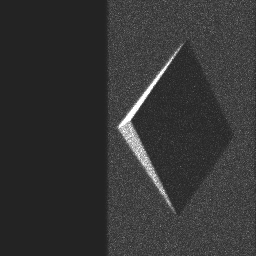}
    
    \includegraphics[width=\figsize\linewidth]{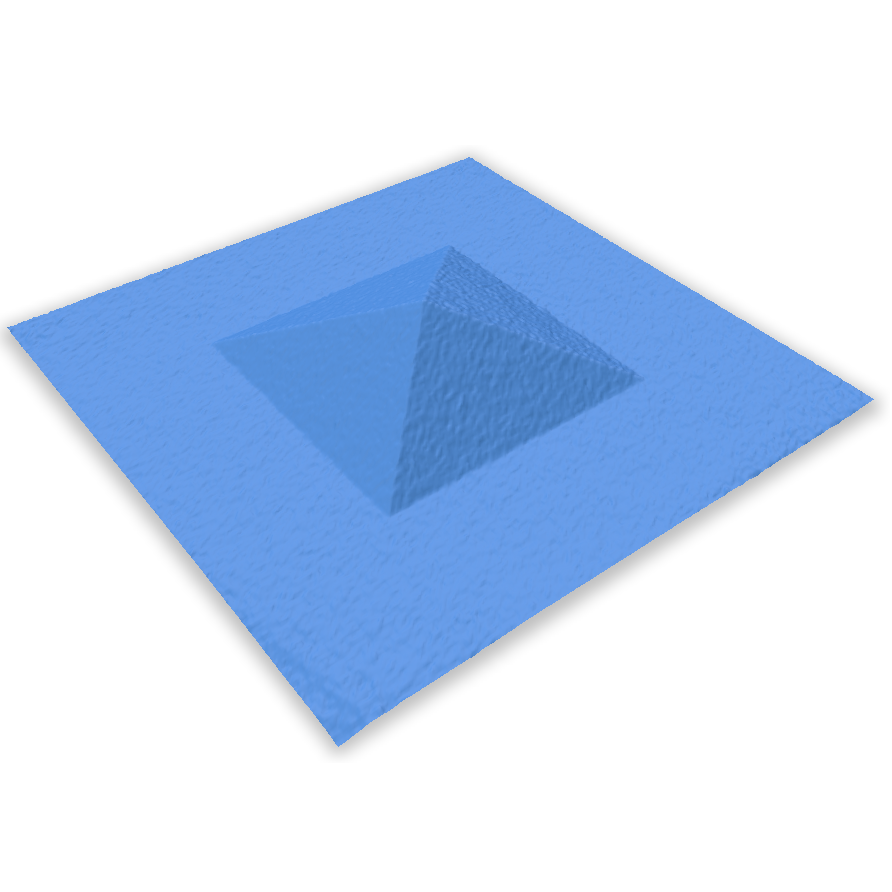}
    \includegraphics[width=\figsize\linewidth]{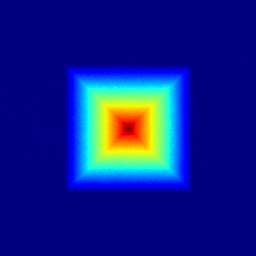}
    \includegraphics[width=\figsize\linewidth]{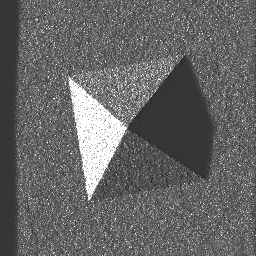}
    \includegraphics[width=\figsize\linewidth]{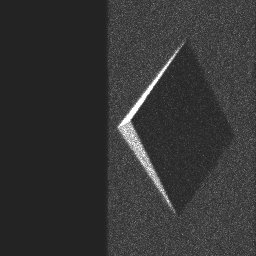}
    
    \includegraphics[width=\figsize\linewidth]{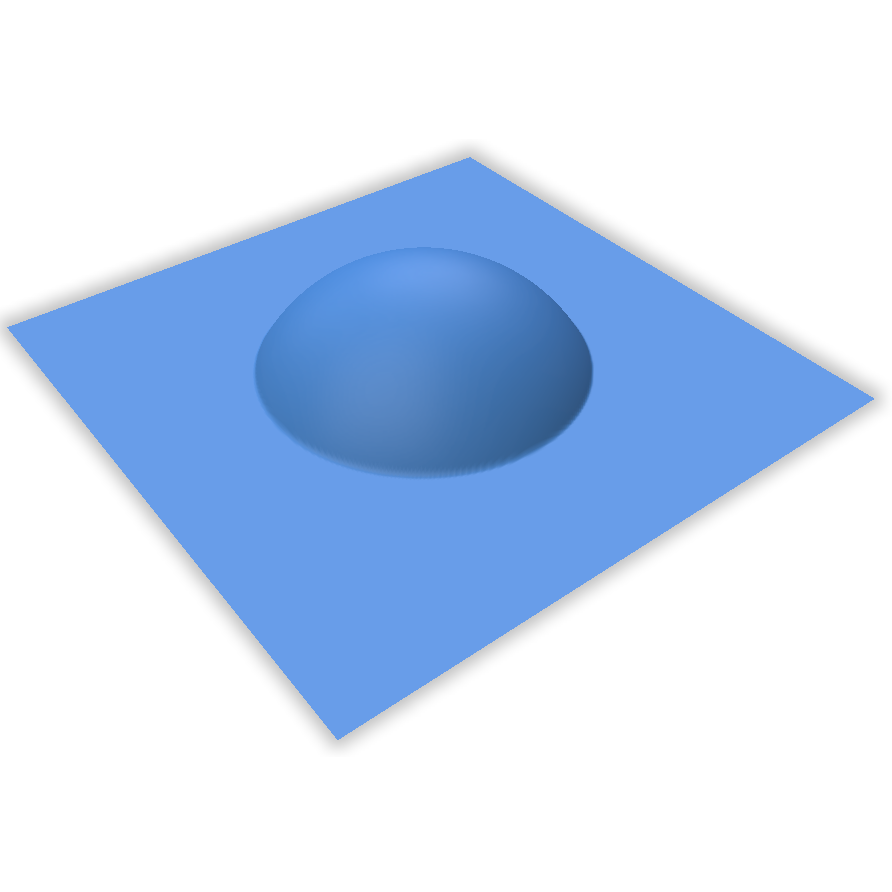}
    \includegraphics[width=\figsize\linewidth]{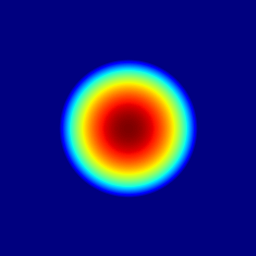}
    \includegraphics[width=\figsize\linewidth]{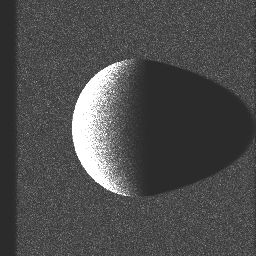}
    \includegraphics[width=\figsize\linewidth]{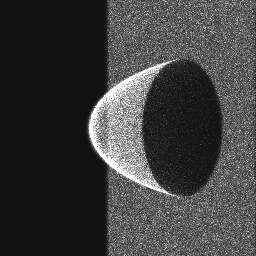}
    
    \includegraphics[width=\figsize\linewidth]{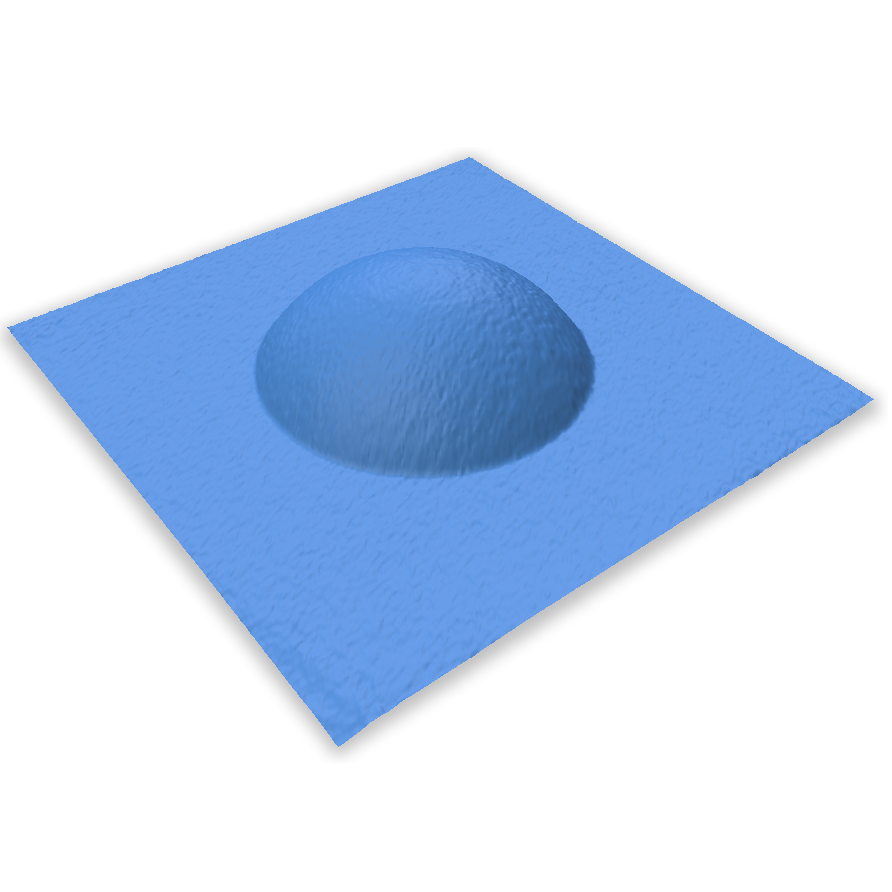}
    \includegraphics[width=\figsize\linewidth]{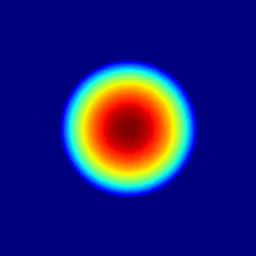}
    \includegraphics[width=\figsize\linewidth]{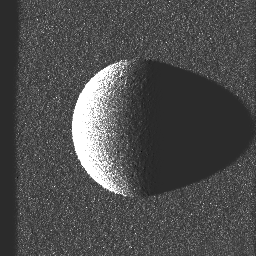}
    \includegraphics[width=\figsize\linewidth]{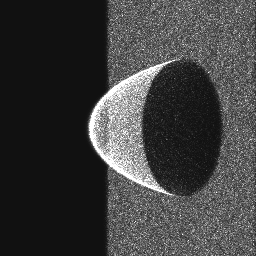}
    \caption{Results of surface learning using five SAR images on two toy examples. The first represents a pyramid (first row is the ground truth and second row the learned model) and the second a round pile (ground truth in the third row and learned model in the last row). From left to right: visualization of the surface model, DSM corresponding to the surface model, and two images corresponding to two views used during training.}
    \label{fig:toy_examples}
\end{figure*}

\begin{figure*}
    \centering
    \includegraphics[width=\figsize\linewidth]{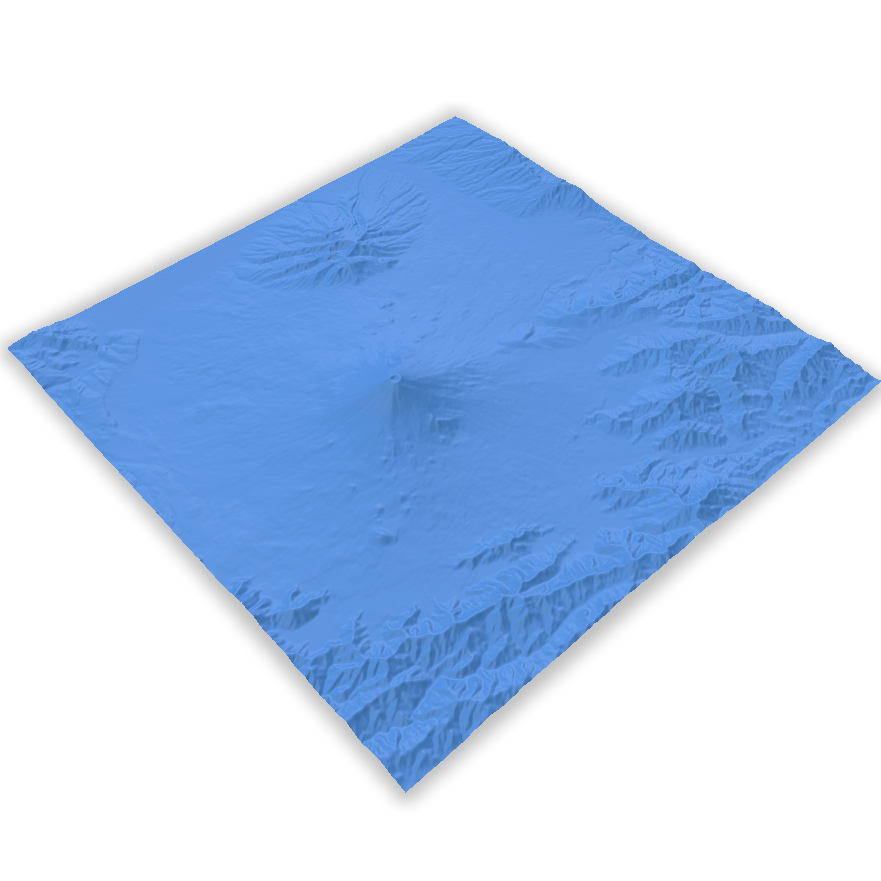}
    \includegraphics[width=\figsize\linewidth]{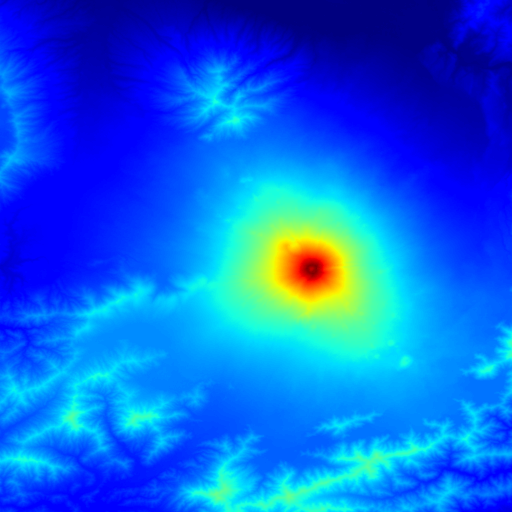}
    \includegraphics[width=\figsize\linewidth]{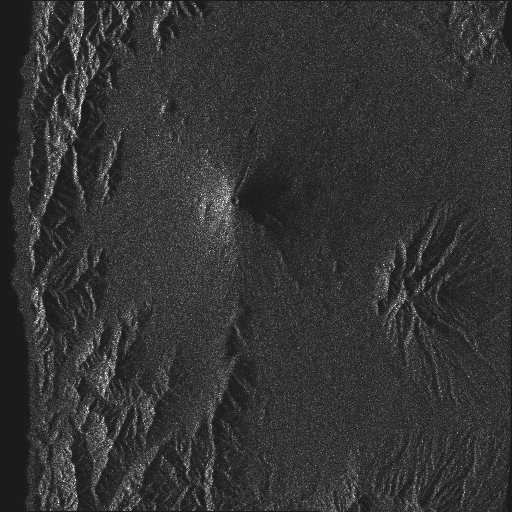}
    \includegraphics[width=\figsize\linewidth]{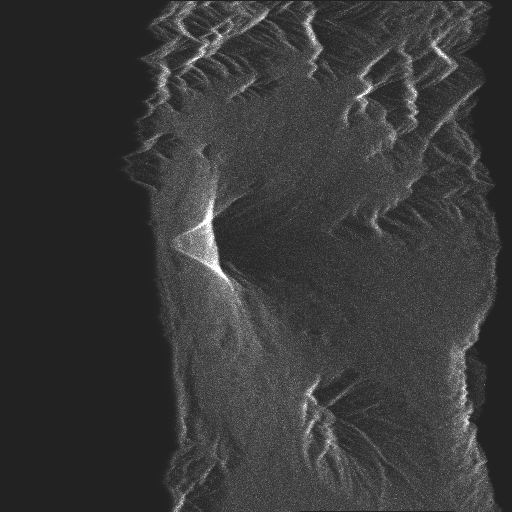}
    
    \includegraphics[width=\figsize\linewidth]{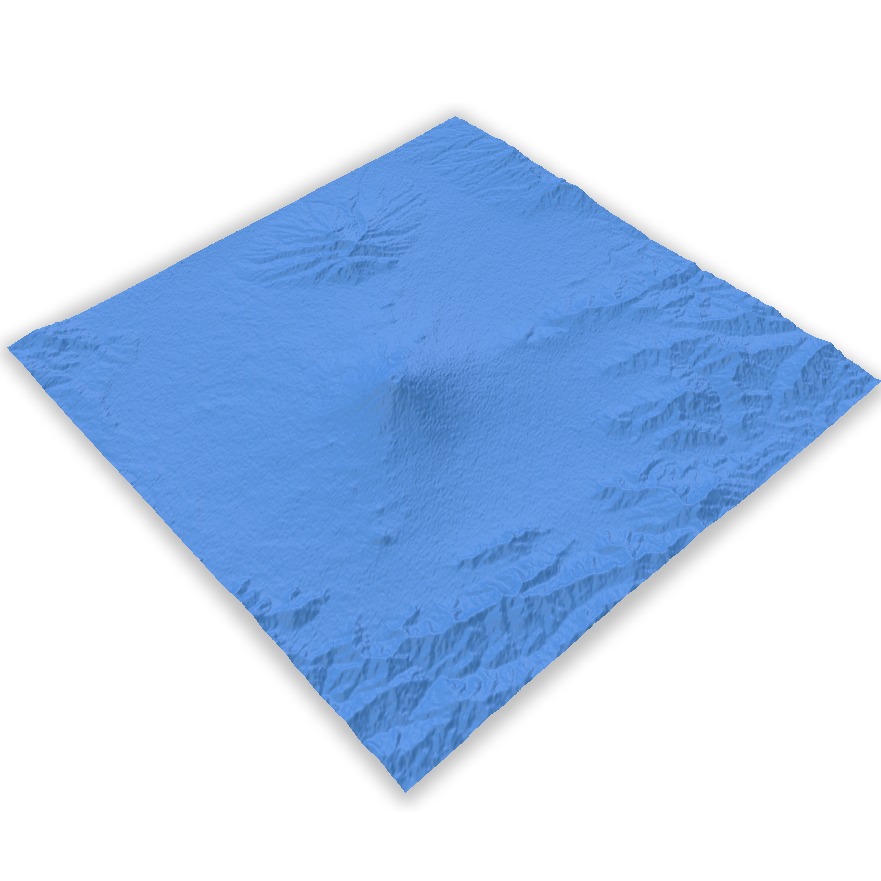}
    \includegraphics[width=\figsize\linewidth]{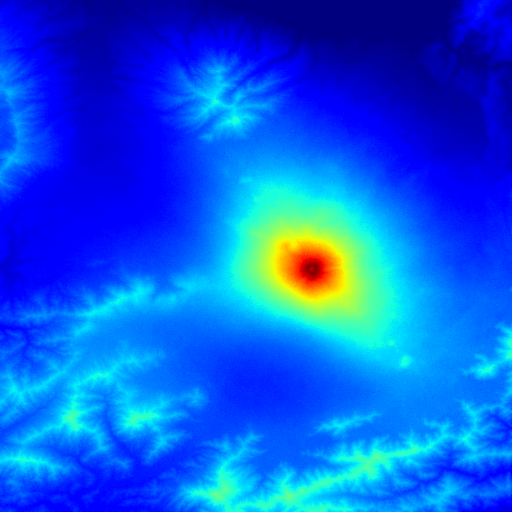}
    \includegraphics[width=\figsize\linewidth]{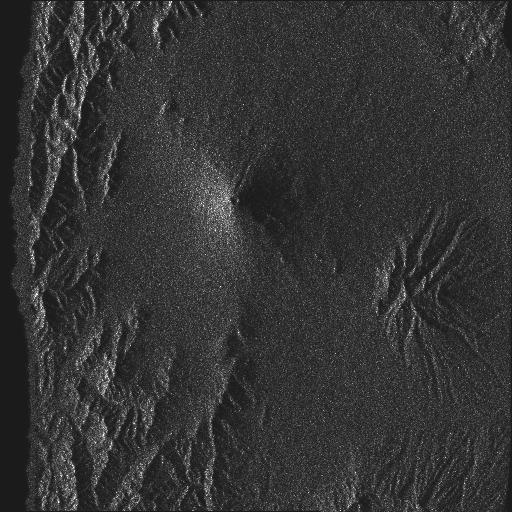}
    \includegraphics[width=\figsize\linewidth]{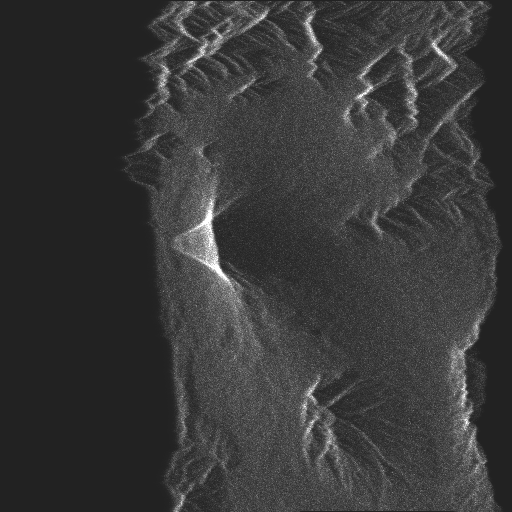}
    
    \includegraphics[width=\figsize\linewidth]{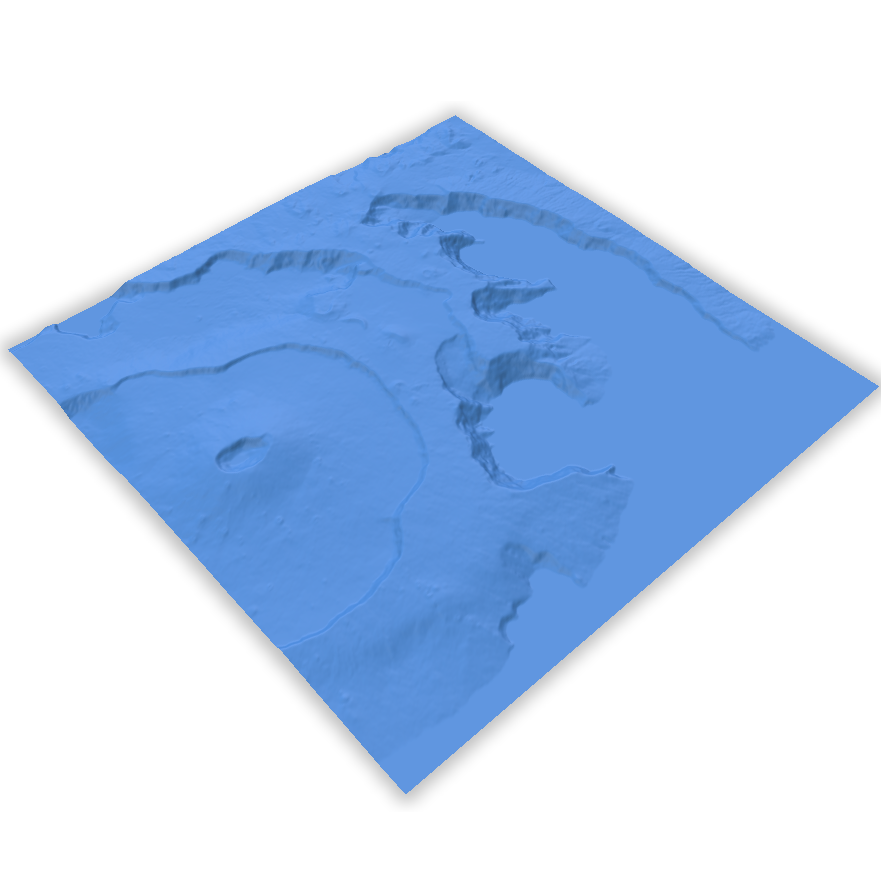}
    \includegraphics[width=\figsize\linewidth]{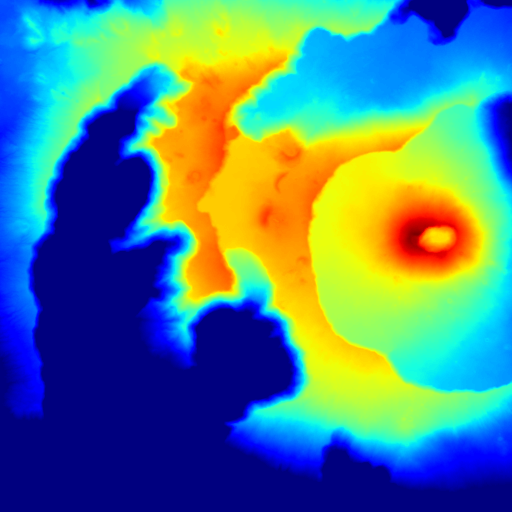}
    \includegraphics[width=\figsize\linewidth]{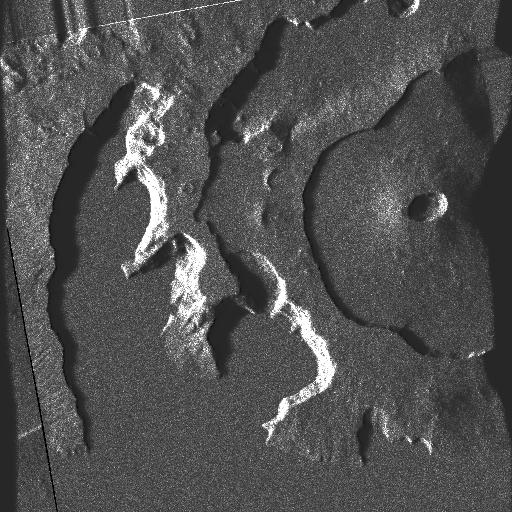}
    \includegraphics[width=\figsize\linewidth]{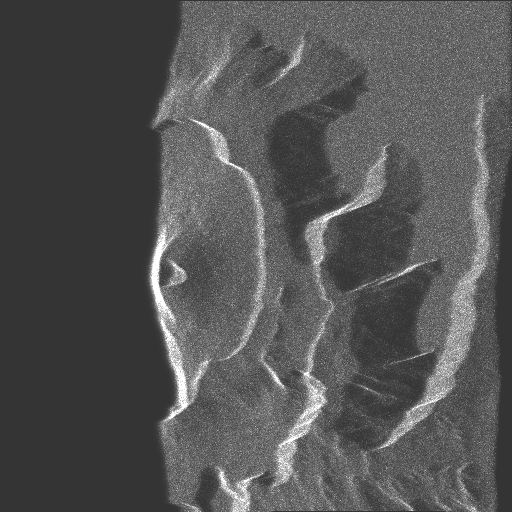}
    
    \includegraphics[width=\figsize\linewidth]{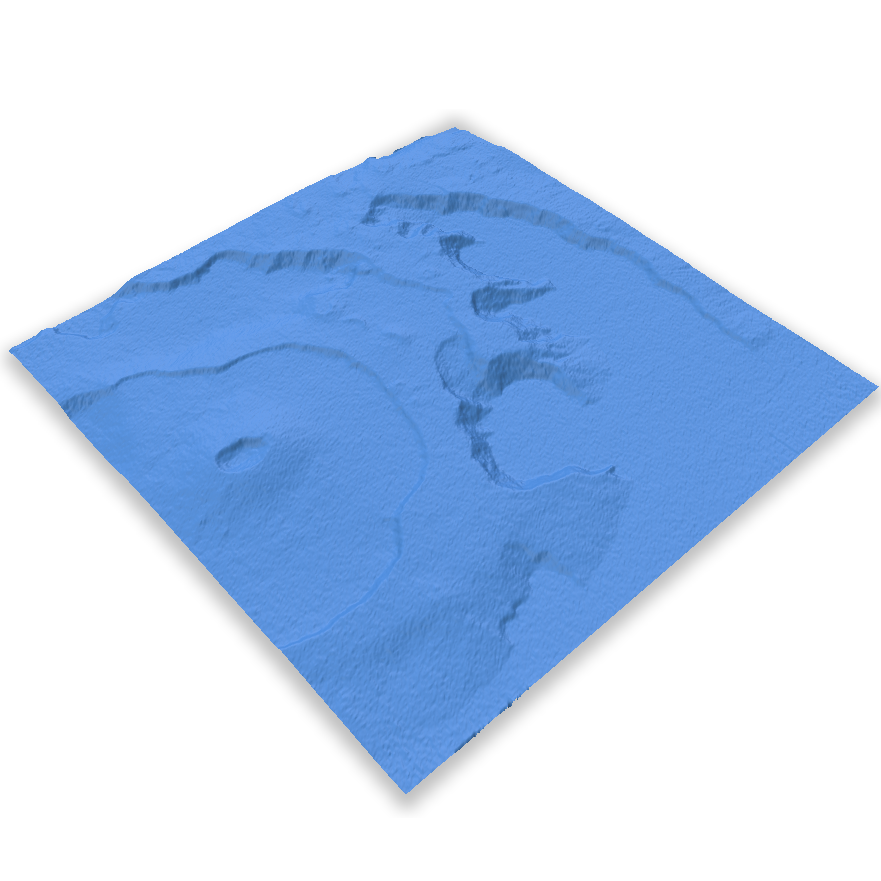}
    \includegraphics[width=\figsize\linewidth]{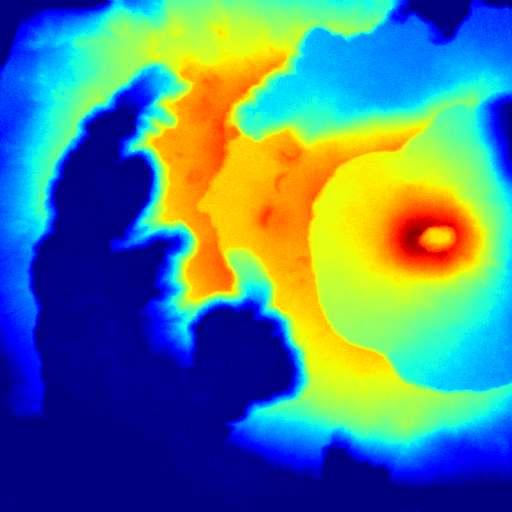}
    \includegraphics[width=\figsize\linewidth]{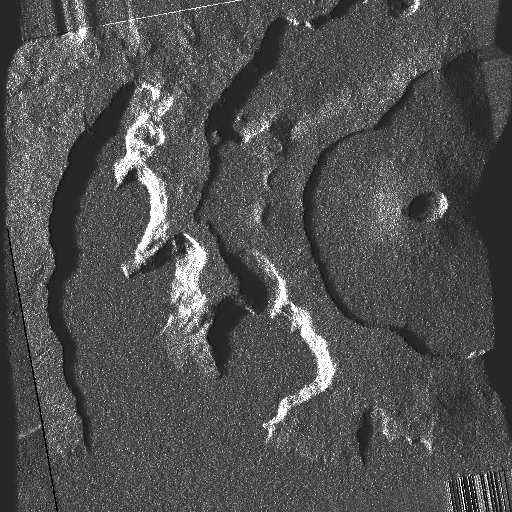}
    \includegraphics[width=\figsize\linewidth]{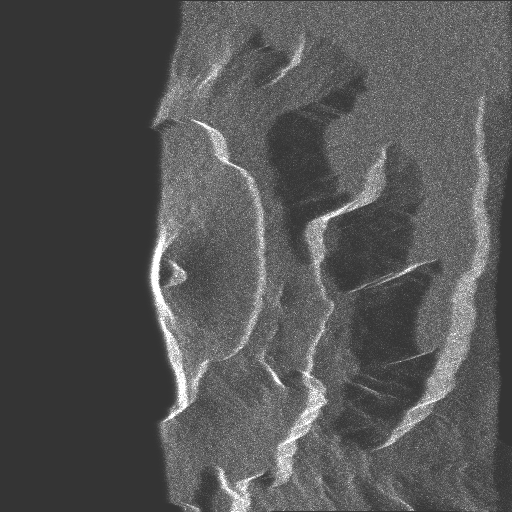}
    \caption{Results of surface learning using five SAR images on simulated data from two real DSMs. The first represents mount Fuji (first row is the ground truth and second row the learned model) and the second the \textit{Piton de la Fournaise} (ground truth in the third row and learned model in the last row). From left to right: visualization of the surface model, DSM corresponding to the surface model, and two images corresponding to two views used during training.}
    \label{fig:real_examples}
\end{figure*}

\paragraph{Surface model implementation.}
In remote sensing, ground surfaces are usually represented by a DSM, \ie a 2D map of the  altitude of the ground at each position. %
{For our experiments we propose to use the same formalism to define the surface in the proposed radar field.} From $dsm: [0,1]^2 \rightarrow [0,1]$, we define 
\begin{equation}
    \mathcal{F}(\bx) = z - dsm(x,y)
\end{equation}
with $\bx = (x,y,z) \in [0,1]^3$. While this function is not exactly an SDF since its value might not necessarily correspond to the exact distance to the surface, it nonetheless defines an implicit surface as in Eq.~\ref{eq:implicit_surface}. The results shown in Sec.~\ref{sec:experiments} were computed using a rasterized DSM that can be assimilated to an image whose pixel represent the height value of the surface. It is the value of each pixel that is then directly optimized. Subpixel heights are estimated using a bilinear interpolation of the nearest values and normals to the defined surface are computed accordingly. One of the advantage is that the Eikonal loss~\eqref{eq:eikonal_loss} is not necessary and it is possible define simple surface regulation terms like 
\begin{equation}
    \sum_{\bx} \sum_{\bm{y} \in \mathcal{N}(\bx)} \|dsm(\bx) - dsm(\bm{y})\|_2^2,
\end{equation}
with $\mathcal{N}(\bx)$ the set of neighbors of $\bx$, that we use for the experiments. Note that the presented concept should remain compatible with the more classic definitions of radiance fields.

\section{Experiments}
\label{sec:experiments}
We apply the proposed radar field model to synthetic examples generated from toy surfaces and real DSMs to demonstrate its potential.

\paragraph{Synthetic data generation.}
{We generate the synthetic SAR acquisitions, using the same image formation model described in Section~\ref{sec:radar_fields}.} Indeed, given that the ground truth is also represented 
{by a DSM, the radar field synthesis equations can be used to simulate}  the images corresponding to the query views. The main difference is that noise needs to be added at the end of the process. %
For this, following Eq.~\eqref{eq:simplified_noise}, we use instead $\tilde{s}(d) = n \times s(d)$ with $n \sim \Gamma(1,1)$ and $s(d)$ defined by Eq.~\eqref{eq:radar_integration}. The optimization process requires only   %
{the generated noisy images and the geometric parameters associated to each view.}

\begin{figure}

\begin{tabular}{@{}l@{}l}
    \hspace{.5em}\includegraphics[width=0.398\linewidth]{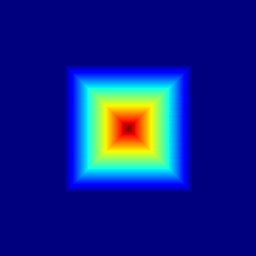}&
    \includegraphics[width=0.398\linewidth]{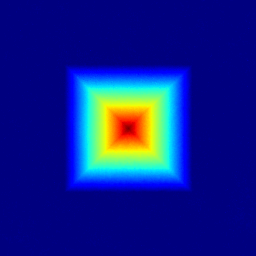} \\
    \hspace{.5em}\includegraphics[trim={7 0 0 0},clip, width=0.49\linewidth]{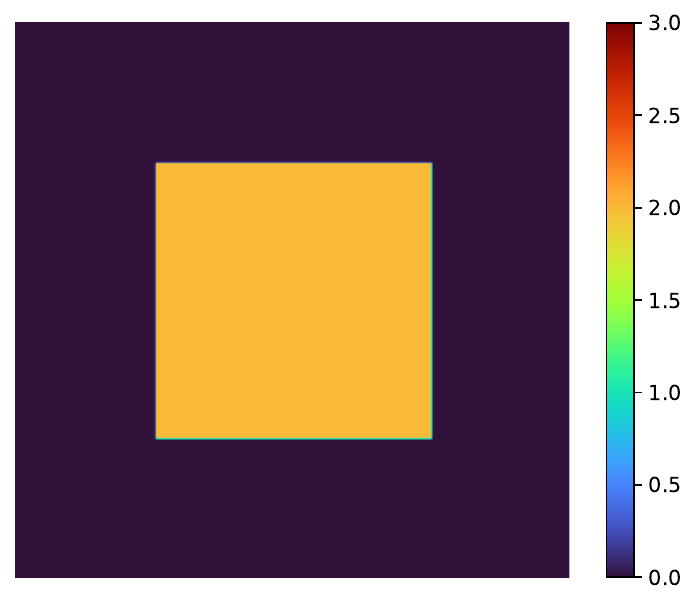}&
    \includegraphics[trim={7 0 0 0},clip, width=0.49\linewidth]{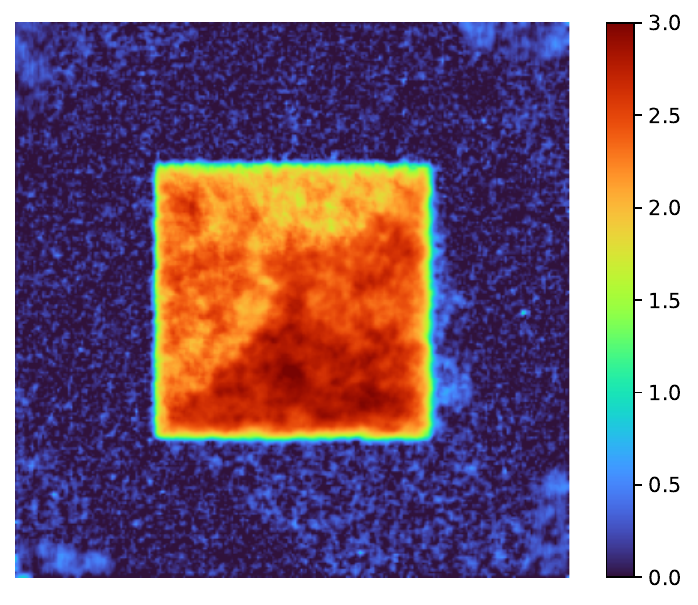}
\end{tabular}
\vspace{-1em}
    \caption{Example showing that it is possible to learn the $\theta$ coefficient (see Eq.\eqref{eq:lambertian_reflectance}) characterizing the surface property (bottom) at the same time as the surface model (top). Left is the ground truth while right is the learned model. The learned model was initialized with $\theta=1$ everywhere.}
    \label{fig:example_specularity_learning}
\end{figure}

\paragraph{Parameters.}
All the experiments presented in this section were done using the same parameters and the images generated corresponds to the same five camera positions.
The models were trained during 10000 steps with an initial learning rate of 1. We reduced the learning rate to 0.1 at epoch 5000 and to $10^{-2}$ at step 8000. Azimuth planes were sampled uniformly from all images and grouped into batches of size 64, and we sampled 256 rays per azimuth plane. Except when specified otherwise, we consider Lambertian surfaces with $\theta=1$.

\paragraph{Results.} Figures~\ref{fig:toy_examples} and \ref{fig:real_examples} show results both with toy examples (a pyramid and a circular pile) as well as simulations using real DSMs (of the mount Fuji and the \textit{Piton de la Fournaise}). These results show that it is indeed possible to learn the ground surface from a few opportunistic SAR acquisitions using the proposed radar fields. By quantifying the altitude reconstruction errors relative to the ground pixel resolution, we observe errors in the order of $10^{-2}$ pixels.
Lastly, Figure~\ref{fig:example_specularity_learning} shows an example of reconstruction of a non Lambertian surface, where the specularity map and geometry are both estimated during the optimization. 
In all these examples, the learned surfaces are very accurate despite the presence of a strong noise in the training data. 
Additional results are available in the supplementary material.

\section{Discussion and conclusion}
\label{sec:discussion}

In this section, we discuss the limitations and possible improvements of the model presented in Section~\ref{sec:radar_fields}. 

\begin{figure}
    \centering
    \includegraphics[width=\linewidth]{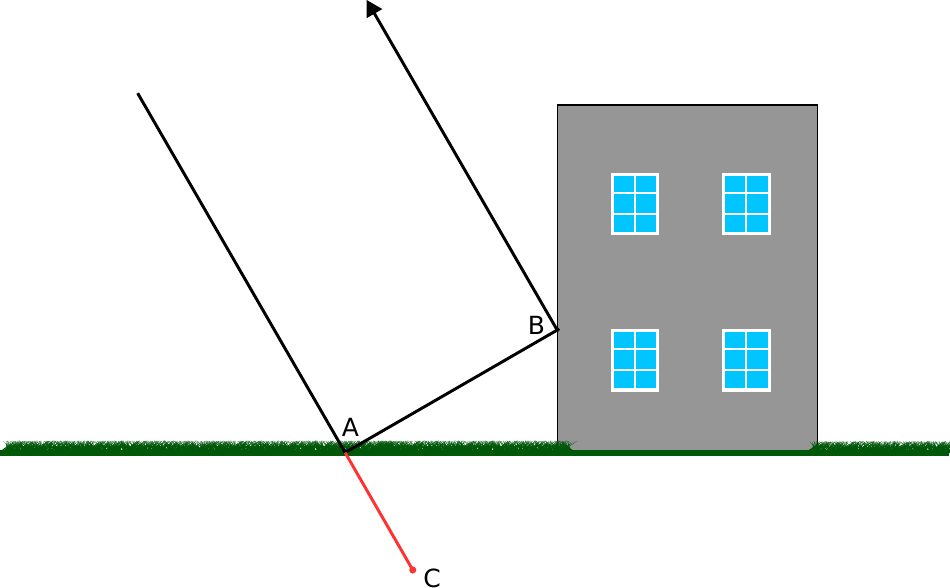}
    \caption{Example of multiple bounces for SAR signal. The signal is partially reflected in $A$ but also bounces and is reflected a second time in $B$. This second bounce is sensed as a virtual reflection in $C$.}
    \label{fig:double_bounce}
\end{figure}

\paragraph{Multi bounce.} Depending on the surface configuration, the SAR signal may be subject to multiple bounces. This phenomena occurs when the signal received by the sensor is not the %
{one directly reflected by an ideal Lambertian surface, but one coming from a second or further reflection}. 
{Corners are a common} source of double bounces. We {distinguish between} two cases: the first is when bounces occur in a given azimuth plane, while the second is a more generic case where bounces are not restricted to that plane.

Figure~\ref{fig:double_bounce} illustrates the case of double bounce in a given azimuth plane. Part of the signal is reflected at the first contact with the surface -- as modeled so far -- but also bounces to another part of the surface before being reflected back to the sensor. This process could be simulated by continuing rays after the first surface intersection, similar to how additional rays are computed in~\cite{mari2023multi} to predict shadows. This increases the complexity of the model by a factor equal to the maximum number of bounces considered. 
{It is important to note that the apparent position of the reflection associated to the double bounce is not the same as the true reflection position. Indeed, the acquisition system measures the traveling time of the impulse signal from emission to reception and as such assumes that the reflection happened at half the traveling time. 
For example, in Figure~\ref{fig:double_bounce}, the signal that bounced from $A$ to $B$ and back to the receiver appears at the same position as if it were reflected by point $C$.} %

The generic case is more complex than the latter because the zero-doppler assumption is not true anymore. This means that a ray corresponding to a given azimuth plane can impact the signal of another azimuth plane. Such a generic multi-bounce scenario would require batching multiple azimuth planes together to render each pixel, making the method highly unfeasible unless strong priors or other constraints are exploited in future work. 

\paragraph{Phase and polarimetric information.} One additional attribute of SAR data that could be incorporated is the phase information. Indeed, SAR images are made of complex numbers containing the intensity and the phase of the signal. SAR Interferometry (InSAR) exploits the phase of multiple images with adequate acquisition configurations. Multitemporal InSAR images can be used to measure small terrain deformation~\cite{massonnet_displacement_1993}, and across-track InSAR \cite{8985551} can be used to build digital surface models (DSM) using a pair of images acquired from two antennas (e.g., SRTM), as mentioned in Section \ref{sec:related_works}. Therefore, we see potential in the phase information to improve altitude accuracy in future work. Similarly, polarimetric information could also be modeled. Polarimetric information is widely used to determine the scattering mechanism~\cite{lee:hal-00351911} and as such can be useful when estimating the surface model. It can also be useful when estimating additional scene properties, such as semantic classification.

\paragraph{Conclusion.} We presented radar fields, an extension of radiance fields to SAR imagery. While the model presented in this paper is a proof of concept tested on simulated synthetic data, we believe that it opens up new exciting opportunities, especially in remote sensing where high-quality SAR images are becoming increasingly common and the generation of surface models from satellite images is a major research topic. Moreover, it also shows a potential to inspire future hybrid models combining both opportunistic optical and SAR acquisitions in a joint framework.

{
    \small
    \bibliographystyle{ieeenat_fullname}
    \bibliography{main}
}

\end{document}